%% file: paper.tex
\documentclass[journal]{vgtc}                     % final (journal style)
\onlineid{1880\textbf{}}

\vgtccategory{Research}

\vgtcpapertype{algorithm/technique}

\title{Rapid and Precise Topological Comparison with \\ Merge Tree Neural Networks}

\author{%
  Yu Qin, Brittany Terese Fasy, Carola Wenk, and Brian Summa}

\authorfooter{
\item
 Yu Qin is with Tulane University. E-mail: yqin2@tulane.edu.
\item
 Brittany Terese Fasy is with Montana State University. E-mail: brittany.fasy@montana.edu.
\item
 Carola Wenk is with Tulane University. E-mail: cwenk@tulane.edu.
\item
 Brian Summa is with Tulane University. E-mail: bsumma@tulane.edu.
}

\abstract{%
    Merge trees are a valuable tool in the scientific visualization of scalar
    fields; however, current
    methods for merge tree comparisons are computationally expensive, primarily
    due to the exhaustive matching between tree nodes. To address this
    challenge, we introduce the Merge Tree Neural Network (MTNN), a learned neural network model designed for merge
    tree comparison. The MTNN enables rapid
    and high-quality similarity computation. We first demonstrate how to train graph
    neural networks, which emerged as effective encoders for graphs,
    in order to produce embeddings of merge trees in vector spaces for
    efficient similarity comparison.  Next, we formulate the novel MTNN
    model that further improves the similarity comparisons by integrating the
    tree and node embeddings with a new topological attention mechanism. We
    demonstrate the effectiveness of our model on real-world data in different
    domains and examine our model's generalizability across various datasets.
    Our experimental analysis demonstrates our approach's superiority in
    accuracy and efficiency. In particular, we speed up the prior
    state-of-the-art by more than $100\times$ on the benchmark datasets while
    maintaining an error rate below $0.1\%$.
}


\teaser{
  \centering
  \includegraphics[width=\linewidth, alt={}]{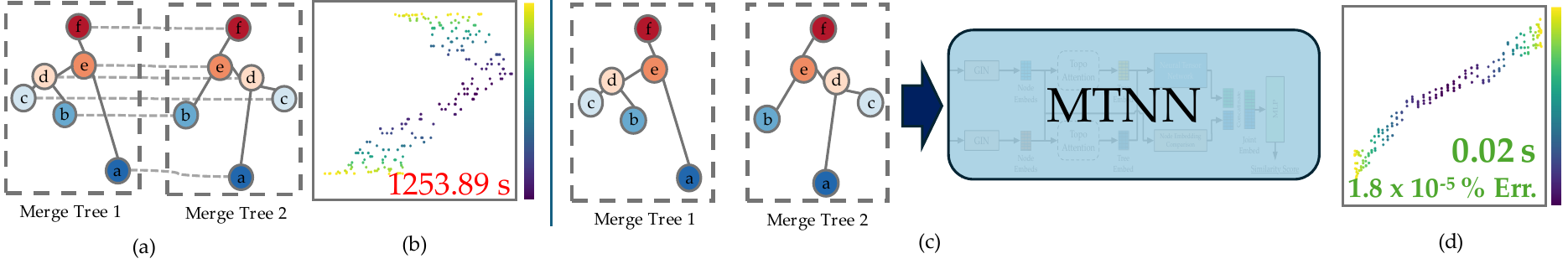}
      \vspace{-18pt}
  \caption{%
            (a) Merge tree distances, where trees are matched and edited, are computationally heavy and slow to compute. This leads to analyses
            that can be a bottleneck in scientific workflows. (b) Here, we show a Multidimensional scaling (MDS) visualization of a test dataset and the significant runtime necessary to produce a pair-wise distance matrix. (c) Rather than
            directly compute tree distances, we treat merge tree comparison as a
            learning task using our novel Merge Tree Neural Network (MTNN) to calculate the similarity. (d) MDS of the same data as (b) using MTNN. This
            reduces comparison times by orders of magnitude (5, here) with extremely low
            error added compared to the state-of-the-art merge tree distance.
  }
  \label{fig:teaser}
}

\graphicspath{{figs/}{figures/}{pictures/}{images/}{./}} % where to search for the images

\usepackage{tabu}                      % only used for the table example
\usepackage{booktabs}                  % only used for the table example
\usepackage{lipsum}                    % used to generate placeholder text
\usepackage{mwe}                       % used to generate placeholder figures
\usepackage[normalem]{ulem}
\useunder{\uline}{\ul}{}
\usepackage{tabularx}
\usepackage{booktabs}

\usepackage{mathptmx}                  % use matching math font
\usepackage{amssymb}
\usepackage{amsmath}

\usepackage{cool}

\newcommand{\T}{\mathcal{T}}

\begin{document}

%%%%%%%%%%%%%%%%%%%%%%%%%%%%%%%%%%%%%%%%%%%%%%%%%%%%%%%%%%%%%%%%
%%%%%%%%%%%%%%%%%%%%%% START OF THE PAPER %%%%%%%%%%%%%%%%%%%%%%
%%%%%%%%%%%%%%%%%%%%%%%%%%%%%%%%%%%%%%%%%%%%%%%%%%%%%%%%%%%%%%%%

\firstsection{Introduction}

\maketitle

\input{body/intro}

\input{body/related}

\input{body/methods}

\input{body/results}

\input{body/conclusion}

\input{body/supplemental}

%\section*{Figure Credits}
%\label{sec:figure_credits}

\section*{Acknowledgements}
This work was supported by NIH R01GM143789, DOE ASCR DE-SC0022873, NSF CCF 2107434, and NSF CCF
2046730.

\bibliographystyle{abbrv-doi-hyperref-narrow}
%\bibliographystyle{abbrv-doi}
%\bibliographystyle{abbrv-doi-narrow}
%\newpage
\bibliography{paper}

%\appendix % You can use the `hideappendix` class option to skip everything after \appendix
%\input{body/appendix}

\end{document}

%% file: body/intro.tex
A fundamental challenge in scientific visualization is analyzing and visualizing
data, such as scalar fields, at speeds that support real-time
exploration. Consider Topological Data Analysis
(TDA)~\cite{edelsbrunner2010computational}, which is now a critical component in
many visualization pipelines due to its ability to extract and succinctly
summarize structural information from complex datasets. TDA has been shown to
effectively visualize~\cite{bremer2010interactive,gyulassy2014stability} and
analyze a wide range of applications in
chemistry~\cite{gunther2014characterizing,bhatia2018topoms},
neuroscience~\cite{maljovec2016topology}, social
networks~\cite{almgren2017extracting}, material
science~\cite{gyulassy2014stability}, energy~\cite{qin2023topological}, and
medical domains~\cite{lawson2019persistent,meng2020weighted,xia2014persistent},
to name a few. However, TDA approaches can be quite slow to compute due to their
computational expense. Therefore, they are difficult to use in scenarios where a
quick answer is needed for an analysis.

In particular, \textit{Merge Trees}
(MTs)~\cite{beketayev2014measuring,edelsbrunner2010computational} serve as
expressive topological descriptors that capture the complex structure of data,
encoding the evolution of topological features with a history that records how
their components merge. Despite being a powerful tool in various domains,
the computational cost associated with distance calculations for \mts has been a
significant limitation. This is due to the core operation of computing the
distance on \mts \remove{needs} \add{needing} to compare the optimal matching
between the nodes of the \mts.
This matching is known to be
NP-hard\cite{agarwal2018computing,bollen2022computing}.
Given the importance, yet great difficulty, of computing \mt
distances, a variety of metrics and pseudo-metrics have been proposed, such as functional distortion
distance~\cite{beketayev2014measuring}, edit
distance~\cite{sridharamurthy2018edit}, interleaving
distance~\cite{gasparovic2019intrinsic,morozov2013interleaving}, and distances
based on branch decomposition and matching~\cite{saikia2014extended}. While many
of these distance metrics are computationally challenging, it is worth noting
that they are not universally NP-hard~\cite{beketayev2014measuring}. Some
metrics may have polynomial-time algorithms under specific conditions.
However, these methods usually require complicated design and implementation
based on discrete optimizations. Despite not being exponential, the time
complexity is usually still polynomial or sub-exponential in the number of nodes
in the \mts.

In this work, we take a different approach. To address the challenge of fast
comparisons, we re-frame it as a \textit{learning} task. Once the similarities
of \mts become learnable, we reduce the comparison runtime by
avoiding the need to compute matchings between two MTs. To do so, we first need an
encoder to map \mts to a vector space, and learn this embedding model to
position similar \mts closely and dissimilar ones far apart in this space. We
focus our investigation on graph neural networks (GNNs) as our primary encoder,
a technique extensively studied for graph embeddings yet unexplored for \mts.

In the following, we introduce the first merge tree neural network (MTNN)
that uses GNNs to learn \mt similarity. In particular, we design a neural network
model that maps a pair of \mts to a similarity score. At the training stage,
the parameters involved in this model are learned by minimizing the
difference between a predicted \remove{similarity scores and a ground truth}
\add{and a true similarity score,}\remove{where
each training data point is a pair of \mts with their true similarity score,} \add{which is the interleaving distance on labeled merge trees~\cite{yan2022geometry} in this work}. At
the test stage, we obtain a predicted similarity score by feeding the
learned model with any pair of unseen \mts.

To effectively learn both the structural and topological information of \mts, we
need to design the neural network architecture carefully. Traditional GNNs
only encode structural information of graphs, not \mts. We introduce two key
strategies to address this gap: firstly, we employ the graph isomorphism
network~(GIN) as our encoder, which excels in identifying structural distinctions in \mts
due to its design for graph isomorphism tasks. Secondly, we develop an innovative
topological attention mechanism designed to identify and highlight
important nodes within a merge tree. This mechanism re-weights nodes based on
their ground truth MT distance and nodes' topological significance (persistence).

We conduct our experiments on five datasets: a synthetic
point cloud, two datasets from time-varying flow simulations, one from a
repeating pattern flow simulation, and one of 3D shapes. On all datasets, MTNN
significantly enhances efficiency, accelerating the runtime by over 100 times
while maintaining a low error rate when compared to a target MT similarity
metric~\cite{yan2022geometry}.

Our contributions can be summarized as follows:
\begin{itemize}
    \vspace{-2pt}
    \item The first neural network model for \mt similarity (MTNN);
    \vspace{-14pt}
    \item A novel topological-based attention mechanism for GNNs;
    \vspace{-4pt}
    \item An evaluation that shows MTNN's extremely precise ($< 0.1\%$ mean
        squared error) and fast ($>100\times$ speedup) similarity measures;
        \vspace{-4pt}
    \item An evaluation of the generalizability of trained models; and
        \vspace{-4pt}
    \item An open-source implementation for
        reproducibility.\footnote{\url{https://osf.io/2n8dy}}
\end{itemize}

The paper is organized as follows: \secref{related} discusses related work,
focusing on the computation of distance between \mts and the application of
learning methods for graph similarity. \secref{pre} provides the
necessary preliminary background. In \secref{gnnmt}, we examine the
using GNNs for \mts. Our proposed model, MTNN, is detailed in
 \secref{mtnn}. \secref{results} presents our results, and
 \secref{conclusion} concludes.

%% file: body/related.tex
\section{Related Work}
\label{sec:related}

Our approach is informed by two main areas of research: (1) the computation of
distances between \mts and (2) learning graph similarity using
graph neural networks (GNNs).
%While some studies have focused on the efficient generation of \mts~\cite{nigmetov2022fast,lukasczyk2023extreem}, our primary interest lies in accelerating \mt comparison.
\add{This section discusses the related works on merge tree distance and learning graph similarity using GNNs. For the definitions of merge trees, their distances, and GNNs, please refer to \secref{pre}.}
\subsection{Merge Trees Distance}
\label{subsec:mtdis}

Many distances between \mts exist:
the interleaving distance~\cite{morozov2013interleaving,gasparovic2019intrinsic}, the edit
distance~\cite{di2016edit,bauer2021reeb}, the functional
distortion distance~\cite{beketayev2014measuring}, and the universal distance~\cite{bauer2014measuring}. While these distances are valuable for their
stability and
discriminatory capabilities, computing these distances often becomes impractical
due to their NP-hard nature~\cite{agarwal2018computing,bollen2022computing}.

Recent efforts aim to define more computationally feasible distances for \mts.
For instance, the edit distance on \mts, as discussed
in~\cite{sridharamurthy2018edit,sridharamurthy2021comparative}, identifies the
optimal edit operations between \mts and has been experimentally shown to be more
discriminative than both the bottleneck and one-Wasserstein~\cite{edelsbrunner2010computational} distances. \add{In the~\cite{sridharamurthy2021comparative}, they} introduced the local merge tree edit distance, specifically designed
to analyze local similarities within \mts. Despite the work
in~\cite{wetzels2023taming} demonstrating greater discriminative power than the
previously proposed edit distances for \mts, by employing a new set of improved
edit operations, the computational cost is significantly higher, which hinders
their practical application.

The interleaving distance requires an initial labeling of the
\mts, followed by the identification of the optimal matching between the labeled merge
trees. This process, which establishes a computable metric known as the labeled
interleaving distance, is introduced in~\cite{gasparovic2019intrinsic}. Yan et
al.~\cite{yan2019structural} adapted this distance for practical use and
incorporated geometric information in the labeling strategy
in~\cite{yan2022geometry}.
Similarly, Curry et al.~\cite{curry2022decorated} employed the
Gromov--Wasserstein distance to label \mts and compute the labeled interleaving
distance.

Branch decomposition trees (BDTs) represent another method that first converts
\mts into BDTs (i.e., transferring edges of \mts as nodes in a new tree), and then
finds pairwise matching between these transformed
trees~\cite{beketayev2014measuring,saikia2014extended,saikia2017global,wetzels2022branch,pont2022principal,pont2023wasserstein}.
This process adds extra computational steps, further increasing complexity.

Applications utilizing \mts distances, such as sketching~\cite{li2023sketching}
or encoding \mts \cite{pont2022principal,pont2023wasserstein}, primarily operate
in the original space, necessitating optimal matching between \mts or their BDT
variants. Machine learning has been used for \mts in the work of Pont et al.~\cite{pont2023wasserstein}, who applied neural
networks to \mts for compression and dimensionality reduction.  Our work focuses
on fast comparison. The previous work also uses a novel but classic auto-encoder with \mts and BDTs
as input, requiring a transformation step.  In addition, their model is not designed to be
generalizable across datasets, and their use of Wasserstein distance in training
is computationally more intensive than our approach.

In summary, existing work either proposes a rigorous definition of distance with
theoretical guarantees but with NP-hard computation or describes a similarity
measure with practical applications that still require optimal matching between
\mts. No existing work focuses on mapping \mts to vector space for efficient
comparison. The closest approach is the work of Qin et al.~\cite{qin2021domain}, who map topological
persistence diagrams to a hash for fast comparison. Merge trees are a more expressive and
more complex topological abstraction than these diagrams.~
In our work, we utilize GNNs to map the \mts to a vector space and
re-frame \mt comparison as a \textit{learning} task.

\subsection{Learning Graph Similarity and Dissimilarity}

Graph (dis)similarity computation is a
fundamental problem in graph theory. The graph edit distance
(GED)~\cite{gao2010survey} is a widely recognized metric between graphs,
defining the minimum number of edit operations required to transform one graph
into another.  Despite its popularity, GED is known to be an NP-hard
problem~\cite{zeng2009comparing}.

With the advancements in graph neural networks (GNNs), it is now possible to
encode graphs into vector spaces
effectively~\cite{kipf2016semi,hamilton2017inductive,xu2018powerful}. This
capability allows GNNs to compute a similarity or dissimilarity score
quickly. These methods employ an end-to-end framework to learn the graph representation
that can, after training, map pairs of graphs to
a similarity~score.

A common approach in this area is the use of a Siamese neural network
architecture~\cite{chicco2021siamese}, which processes each graph independently, but in parallel, to aggregate information. A feature fusion mechanism then captures
the inter-graph similarities, and a multi-layer perceptron (MLP) is applied for
regression analysis. This method is typically trained in a supervised manner
using the mean squared error (MSE) loss against the ground truth similarity
scores.

Many GNN-based approaches for learning graph similarity have great promise due
to the competitive performance in both efficiency and
efficacy~\cite{li2019graph,bai2019simgnn,bai2020learning,qin2021slow}. For
example, the graph matching network (GMN)~\cite{li2019graph} is the first deep
graph similarity model, which computes the similarity between two given graphs
by a cross-graph attention mechanism. SimGNN~\cite{bai2019simgnn} turned the
graph similarity task into a regression task, and leveraged the graph convolutional network (GCN)~\cite{kipf2016semi} layers with
self-attention-based mechanism on the model.

However, despite the growing popularity of GNN-based methods for graph
similarity, their application to topological descriptors like \mts has not been
explored. \add{Topological deep learning is a rapidly evolving field, although it primarily focuses on using topological features to enhance deep learning models~\cite{zia2024topological,papamarkou2024position,hensel2021survey}. Our perspective focuses on the inverse: designing deep learning models specifically for topological descriptors.} Our research establishes the initial connection between GNNs and \mts,
potentially laying the groundwork for future developments in machine learning
and TDA.

%% file: body/methods.tex
\section{Preliminaries}
\label{sec:pre}

In this section, we outline the foundational concepts of this work,
beginning with merge trees induced by scalar fields and the common distances
used for topological comparisons. Then, we describe graph neural networks
(GNNs), which serves as the core architecture for encoding merge trees for topological comparison.

\subsection{Scalar Fields and Merge Trees}

Consider a dataset represented as $(\mathbb{X}, f)$, where $\mathbb{X}$ denotes
a topological space, and $f: \mathbb{X} \rightarrow \mathbb{R}$ denotes a scalar
function, which is a continuous real-valued function. This function $f$ assigns
a real number to each point in $\mathbb{X}$, reflecting a characteristic of
interest within the dataset. This is often referred to as a {\em scalar field}.

An equivalence relation \(\sim_f\) is defined on \(\mathbb{X}\), where \(x
\sim_f y\) if both \(x\) and \(y\) belong to the same connected component of the
sub-level set \(\mathbb{X}_a\) for a threshold value \(a\) in \(f\).
Consequently, two points are equivalent under \(\sim_f\) if they exhibit a
shared characteristic under the threshold \(a\).

\begin{figure}[t]
    \centering % avoid the use of \begin{center}...\end{center} and use \centering instead (more compact)
    \includegraphics[width=\linewidth]{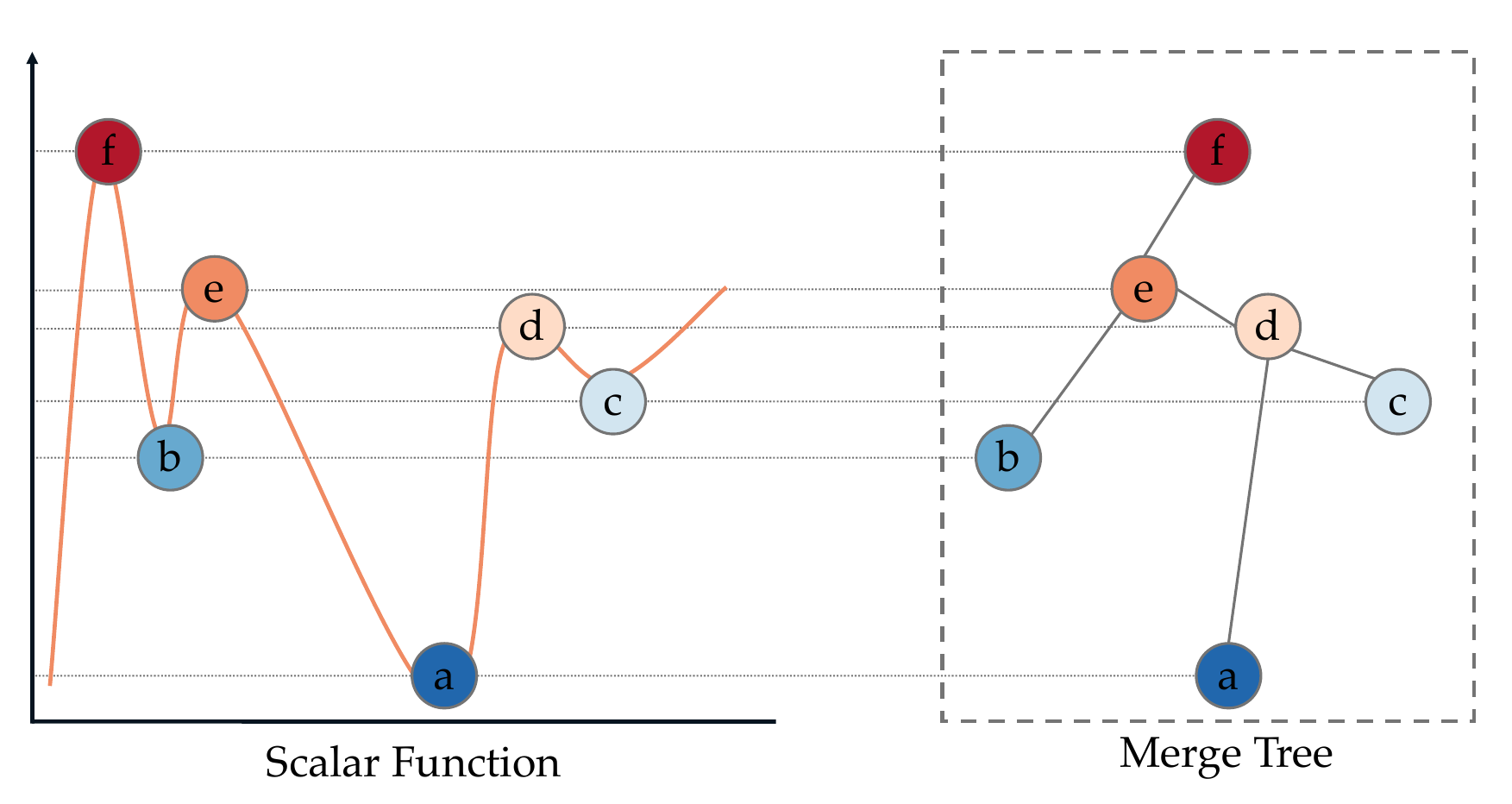}
    \vspace{-18pt}
    \caption{An illustration of merge tree. On the left, a scalar function is represented by an orange line, highlighting the critical points, and showcasing how these points merge and connect topologically. On the right, the corresponding merge tree of sub-level set filtration.
    }
    \label{fig:mt-exp}
    \vspace{-18pt}
\end{figure}

This leads to the definition of \(\mathcal{T}_f^- := (\mathbb{X}_a, f)\) as the
\textit{join tree} for the dataset \((\mathbb{X}, f)\). The \textit{join tree}
encapsulates how dataset components merge as the threshold \(a\) varies.
The \textit{split tree}  \(\mathcal{T}_f^+ := (\mathbb{X}_a, f)\) is similarly
constructed using super-level set to depict component splits with
changing thresholds. Each of these two trees is called a
\textit{merge tree}~\(\mathcal{T}_f := (\mathbb{X}_a, f)\), illustrating the evolution of
topological features in relation to \(f\).
In this work, we use join
trees to capture the connectivity of sub-level sets, with
the global maximum being the tree's root. Example merge tree for a 1D
function is shown in \figref{mt-exp}.

\paragraph{Persistence} Persistence is a quantity derived from persistent
homology~\cite{edelsbrunner2010computational} that tracks the lifetime of a
topological feature. This is often important in analysis. For instance, low
persistence features are often deemed to be noise, while high persistence
features are considered to encode important topological properties. In the
context of our work, persistence is used to track the evolution of connected components of a sub-level set of a scalar function. A persistence pair ($b,d$) represents a topological feature that
is born in the sub-level set $\mathbb{X}_b$ and dies going into the sub-level
set $\mathbb{X}_d$. $b$ and $d$ correspond to the critical points $m$ and $s$
where $b=f(m)$ and $d=f(s)$.  In our join trees, a feature is born at a
minimum and dies when it merges with an older, in terms of~$f$, feature.  The {\em persistence} of such a pair is defined as the difference in function
value at the two critical points, $d - b$.

%The merge tree is a topological abstraction that summarizes the persistence pairs, so that each path from the root to a leaf node traces the evolution of a connected component within the sublevel sets of $f$.

\subsection{Distance on Merge Trees and Graphs}
Below, we first define graph edit distance, which GCNs primarily focus on reproducing.
We then describe how edit distance on merge trees differs.  Finally, we
describe the distance we use as our ground truth, the interleaving distance.
%\paragraph{Wasserstein Distance} Given two persistence diagrams, $D_i$ and $D_j$,
%this distance quantifies the minimal cost to map one diagram to the other. It involves
%optimally pairing points from $D_i$ and $D_j$ and summing the distances between
%matched points, which are calculated using the endpoints of these intervals.

%Formally, the $p$-Wasserstein distance is expressed as:
%\[
%    W_p(D_i, D_j) = \left(\min_{\gamma} \sum_{i} ||b_i - \gamma(b_i)||^p\right)^{1/p},
%\]
%where $\gamma$ denotes the optimal bijection between intervals of $D_i$ and
%$D_j$, and $||b_i - \gamma(b_i)||$ represents the Euclidean distance between the
%matched interval endpoints. This metric effectively captures the similarity
%between two sets of topological features, offering a robust way to compare
%different datasets' topological structures.

\paragraph{Graph Edit Distance}
The graph edit distance (GED)~\cite{gao2010survey} between two graphs $G_1=(V_1,E_1)$ and
$G_2=(V_2,E_2)$ is the minimum cost of transforming $G_1$ into $G_2$ using
node and edge insertions, deletions, and substitutions. It's defined as:
\vspace{-8pt}
\[
    D_e(G_1, G_2) := \min_{S \in O} \left\{ \sum_{op \in S} \delta(op) \right\},
\vspace{-6pt}
\]
where \(O\) represents the set of all valid sequences of graph edit operations
that transform \(G_1\) into \(G_2\), \(S\) is a specific sequence of graph edit
operations from \(O\), and \(\delta(op)\) is the cost function that assigns a
non-negative real number to each edit operation \(op\) in the sequence \(S\).
This operation could be the insertion, deletion, or substitution of a node or
an edge.

Each operation \(op\) has an associated cost, and $\sum_{op \in S} \delta(op)$, the total cost
of a sequence \(S\), is the sum of the costs of its individual operations. The
GED is then determined by finding the sequence \(S\) with the minimum total cost
that transforms \(G_1\) into \(G_2\).

\paragraph{Edit Distance on Merge Trees}
The edit distance between merge trees builds on tree edit distance, which is a
specific case of GED where the cost operation is only on nodes. Given two merge
trees, denoted as~\(\T_1\) and \(\T_2\), it is defined
as~\cite{sridharamurthy2018edit}:
\vspace{-6pt}
\[
    D_e(\T_1, \T_2) := \min_{S \in O} \{\delta(S)\},
\vspace{-6pt}
\]
where \(O\) represents the set of valid tree edit operations, and \(S\)
represents a sequence of these tree edit operations that transform \(\T_1\) into
\(\T_2\). The cost function \(\delta\) assigns a non-negative real number
to each edit operation, which is defined as~\cite{sridharamurthy2018edit}:
\vspace{-6pt}
\[
    \begin{array}{lll}
    \delta(m \rightarrow s)
    = \min \left\{
        \begin{array}{cc}
            \max(|b_m-b_s|,|d_m-d_s|), \\
            \frac{(|d_m-b_m|+|d_s-b_s|)}{2}
        \end{array}
        \right\} \\
    \delta(m \rightarrow \lambda) = \frac{|d_m-b_m|}{2}\\
    \delta(\lambda \rightarrow s) = \frac{|d_s-b_s|}{2}
    \end{array}
\vspace{-6pt}
\]
where $\lambda$ denotes the empty set and $m$ and $s$ are nodes in $\T_1$
and~$\T_2$, respectively. The first cost is for a node relabel
operation from $m$ to $s$. Next is deleting a node $m$. The last cost is adding
a node $s$.

\begin{figure}[h]
    \centering
    \includegraphics[width=\linewidth]{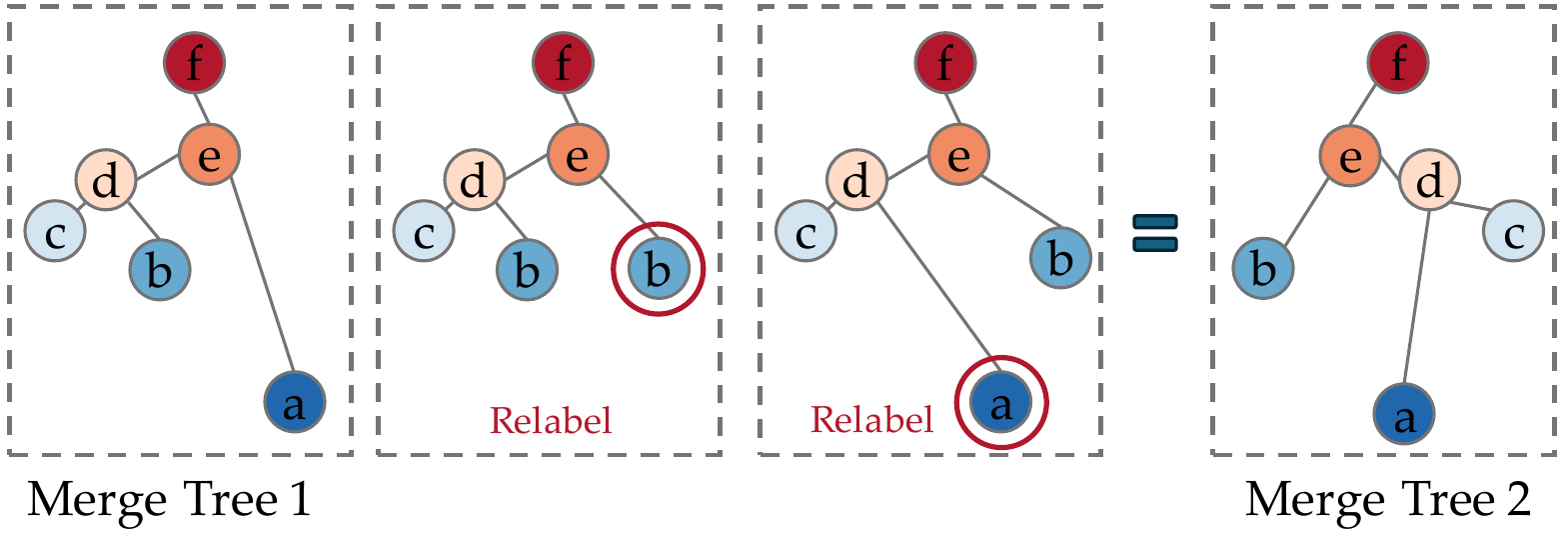} %mtdis}
    \vspace{-18pt}
    \caption{Edit distance on merge trees example. The transformation from merge tree 1 to merge tree 2 needs two nodes to relabel operations, where the relabeled node is highlighted in the red circle.
    }
    \label{fig:mtdis}
    %\vspace{-8pt}
\end{figure}

This is similar to GED, but the cost is formulated to account
for topological persistence.
Consider nodes $m \in \T_1$ and~$s \in \T_2$, each node encodes a topological
feature, $(b_m,d_m)$ and $(b_s,d_s)$. In particular,~$b_m$ is the function value
where $m$ is born. $d_m$ is the function value where $m$ dies.
For sublevel sets, $b_m < d_m$. This distance computation is shown in \figref{mtdis}.

\paragraph{Interleaving Distance on Merge Trees}
To compute the interleaving distance on \mts, the trees need to be labeled first. A labeled \mt, denoted as  \(T = (\T, \pi)\), including a \mt $\T$ with a labeling $\pi:[a] \rightarrow V_{\T}$ where $[a]$ is the set of labels, $\{1, ..., a\}$, and  $V_{\T}$ is the set of \mt vertices~\cite{gasparovic2019intrinsic}. $\pi$ only needs to be surjective since a vertex can have multiple labels. The interleaving distance on labeled \mts is calculated based on the induced matrix $T_M(\T,\pi)$. This matrix also can be referred to as the least common ancestor (LCA) matrix and defined as:
\vspace{-6pt}
\[
    T_M(i,j) = f(LCA(\pi(i),\pi(j)),
\vspace{-6pt}
\]
where $f(LCA(\cdot))$ denotes the function value of LCA of a pair of vertices with labels $i$ and $j$, $1 \leq i,j \leq a$.
Given two labeled \mts $T_1=(\T_1,\pi_1)$ and $T_2=(\T_2,\pi_2)$ that share the same set of labels, $[a]$, the interleaving distance between labeled \mts is defined as:
\vspace{-6pt}
\[
    D_i(T_1,T_2) = \| T_M^1 - T_M^2\|_\infty,
\vspace{-6pt}
\]
where $\|T_M\|=\max_{ij}|T_{Mij}|$ is the $L_\infty$ norm.
In~\cite{yan2022geometry}, they proposed geometry-aware labeling strategies and,
for brevity, we refer the reader to their paper for those details. But, at a
high level, their labeling minimizes a cost function that accounts for the
geometric structure of the tree along with the function value differences
between nodes in the tree. In this way, topological persistence is encoded in
the labeling strategy. The interleaving distance is used as the ground truth \mt distance in our training.

\subsection{Graph Neural Networks (GNNs)}
\begin{figure}[h]
    \centering % avoid the use of \begin{center}...\end{center} and use \centering instead (more compact)
    \includegraphics[width=\linewidth]{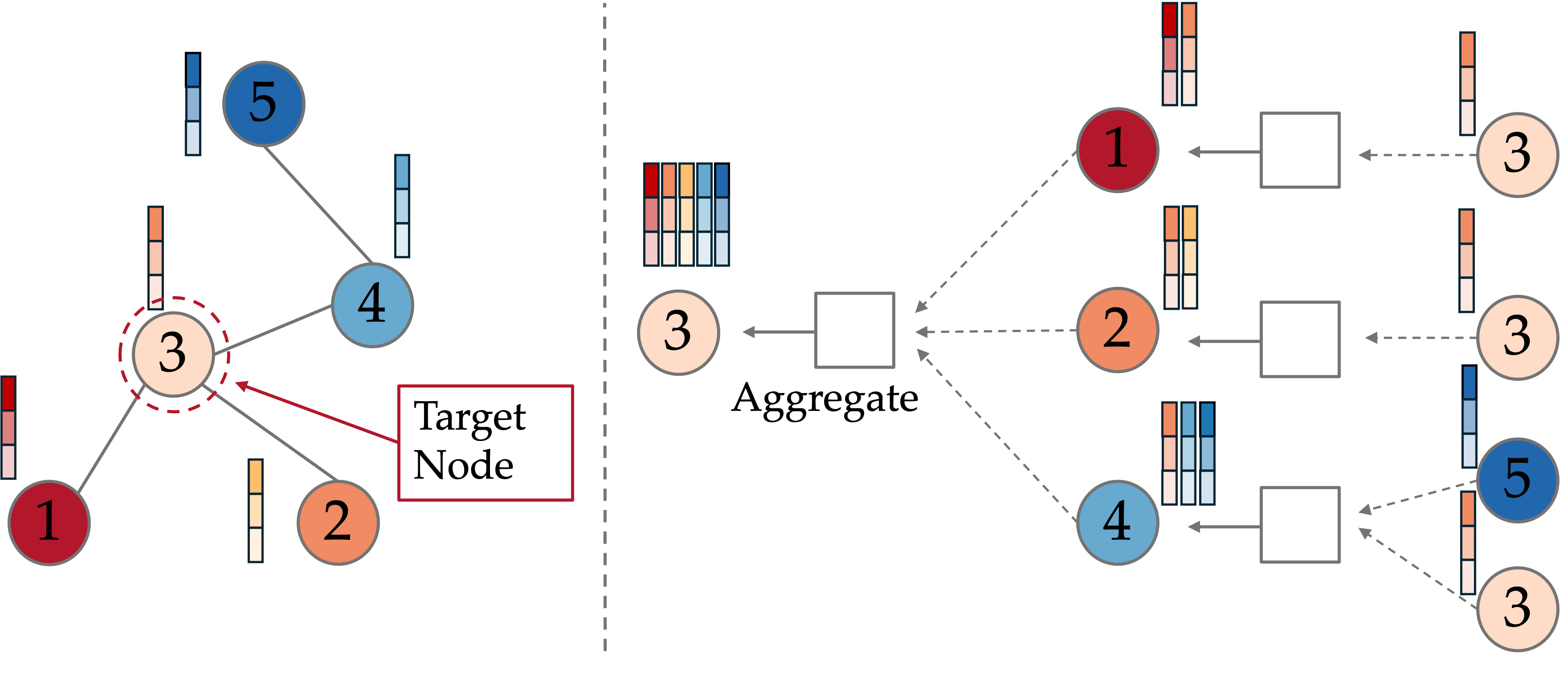} %gnn}
    %\vspace{-18pt}
    \caption{Example of the GNNs process for a single node. Left: a graph with
    node features. Right: message function and update function for the target
    node 3. This node receives information from its directly connected
    neighbors, nodes 1, 2, and 4. Notably, node 4 further communicates a message from its adjacent node 5. In the end, the feature of node 3 is updated by aggregating all incoming messages.
    }
    \label{fig:gnn}
    \vspace{-16pt}
\end{figure}
Given a graph $G = (V, E)$ with nodes $V$ and edges $E$, where each node~$v\in
V$ has an initial feature vector $h_v^{(0)}$, GNNs update the feature
representation of each node by leveraging the structural context provided by the
graph~\cite{hamilton2017inductive}. In particular, each node aggregates features from its neighbors
(\textbf{Message Function}), updates its own features (\textbf{Update
Function}), and finally produces an embedding (\textbf{Embedding}) that
represents either the node's or the entire graph's comprehensive features. As Fig. \ref{fig:gnn} shows, the message function is responsible for aggregating information from a node's neighbors, which is then integrated into the node's feature vectors through the update function.

\textbf{Message Function:}
The message function aggregates features from neighboring nodes. For a node
\(v\), the message function \(M\) can be defined as:
\[
    m^{(l+1)}_v = \sum_{u \in \mathcal{N}(v)} M(h^{(l)}_v, h^{(l)}_u, e_{vu}),
    \vspace{-6pt}
\]
where \(h^{(l)}_v\) is the feature vector of node \(v\) at layer \(l\), \(\mathcal{N}(v)\) denotes the neighbors of \(v\), and \(e_{vu}\) represents the edge features between nodes \(v\) and \(u\).

\textbf{Update Function:}
The update function \(U\) integrates the aggregated messages with the node's
current state to compute its new state:
\vspace{-6pt}
\[
    h^{(l+1)}_v = U(h^{(l)}_v, m^{(l+1)}_v),
\vspace{-6pt}
\]
where \(h^{(l+1)}_v\) is the updated feature vector of node \(v\) at
layer~\(l+1\).

\textbf{Embedding:}
After processing through \(L\) layers of the GNN, the final embedding \(z_v\)
for a node \(v\) is obtained, which can be used for downstream tasks like
classification or clustering:
\vspace{-6pt}
\[
    z_v = h^{(L)}_v.
\vspace{-6pt}
\]
For a whole graph embedding, an aggregation function (e.g., sum, mean, or max) is
applied over all node embeddings to produce a single vector representing the
entire graph. \add{GNNs do not always generalize across different graph sizes, especially from small to large graphs, as noted in~\cite{yehudai2021local}. Therefore, the number of layers is determined by the graph's size and complexity. For large graphs, careful tuning and techniques like hierarchical pooling or attention mechanisms are essential for scalability.}
Through these mechanisms, GNNs offer a powerful framework for
learning from graph-structured data, which we adapt to the context of merge
trees for our topological comparison.

\section{GNNs Meet Merge Trees}
\label{sec:gnnmt}

\begin{figure}[h]
    \vspace{-8pt}
    \centering % avoid the use of \begin{center}...\end{center} and use \centering instead (more compact)
    \includegraphics[width=\linewidth]{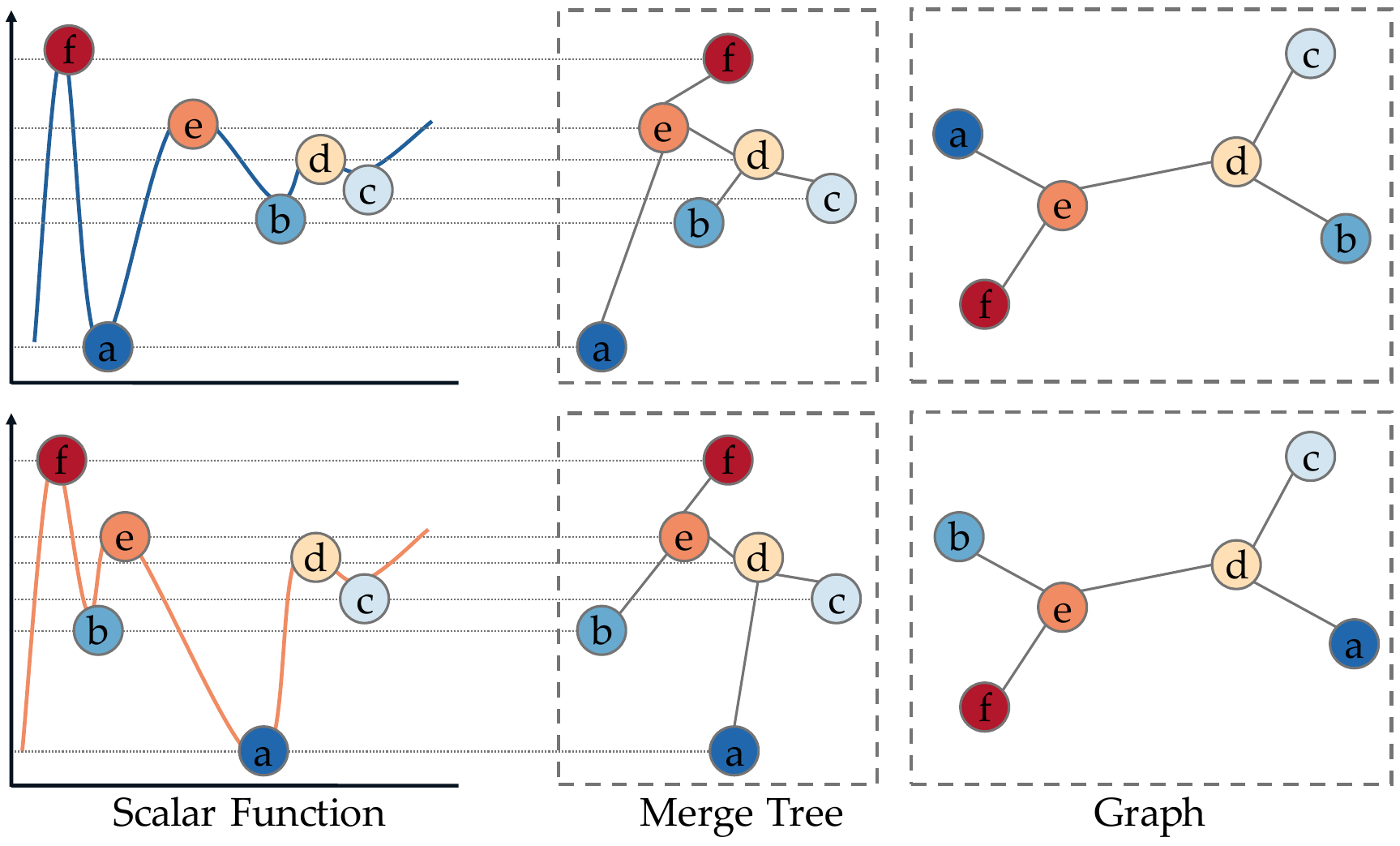}
    \vspace{-18pt}
    \caption{An illustration of two merge trees and their graph structure. From
        left to right: critical points (dots) in two scalar functions, showcasing
        how these points merge and connect topologically. In the middle are
        corresponding merge trees, highlighting structural differences. Finally, the
        graph structure of the two merge trees is identical but differs in function value labels.}
    \label{fig:mt-graph}
    \vspace{-8pt}
\end{figure}

This section describes our methodology for adapting GNNs to \mts. While GNNs have
demonstrated effectiveness in learning on graph-structured data, the
application to \mts is not directly translatable due to unique topological
characteristics inherent to \mts. See \figref{mt-graph}.
To address this, we enrich the node features with topological
information. Specifically, we assign an attribute to each node derived from its
scalar function value. Moreover, our objective goes beyond learning the
structure of \mts; we are focused on a more complicated task, learning \mt
distance. To this end, we employ GCN similarity~\cite{bai2019simgnn}, a model
initially designed for graph similarity learning, adapting it to the context of
\mts. This adaptation allows us to establish a baseline for our approach. More
details are given below.

\begin{figure*}[h]
    \centering % avoid the use of \begin{center}...\end{center} and use \centering instead (more compact)
    \includegraphics[width=\linewidth]{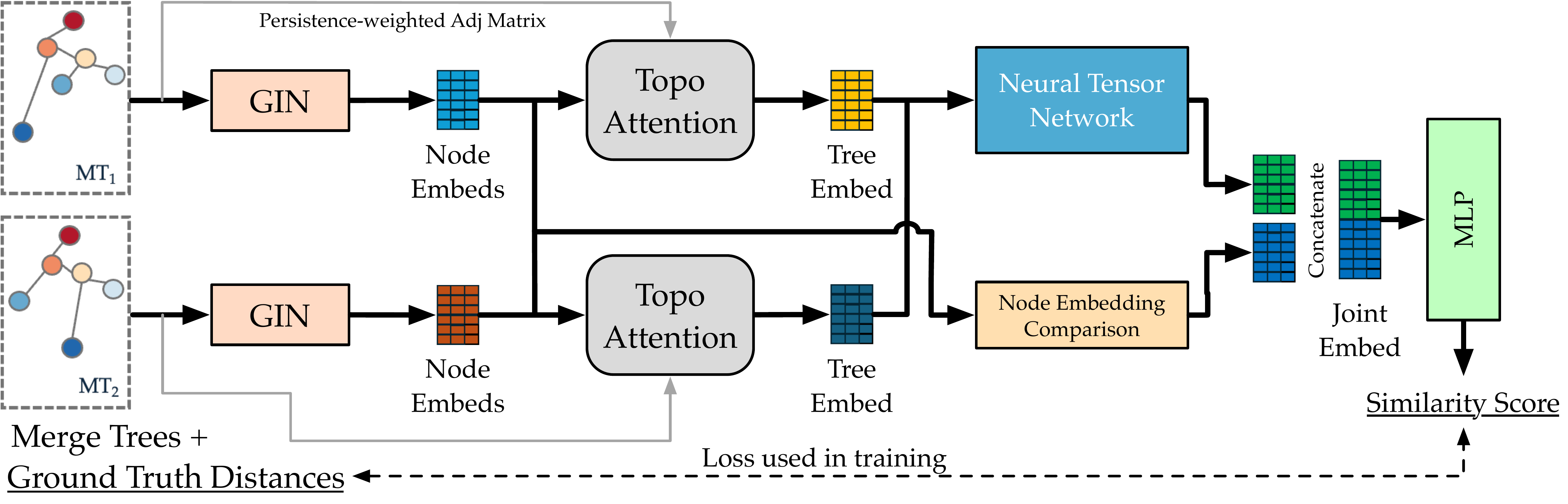}
    \vspace{-18pt}
    \caption{The architecture of our merge tree neural network (MTNN) operating on a pair of \mts ($MT_{1,2}$). First, adjacency matrices for each \mt with node feature are weighted by function value and are fed into a GIN. This produces node embeddings for both.  Next, our topological attention produces tree embeddings from the node embeddings. A persistence-weighted adjacency matrix is an additional input for this step. The node embeddings are then compared, and tree embeddings are fed into a neural tensor network. These outputs are concatenated to form a joint embedding. This joint embedding is fed into a multi-layered perceptron (MLP) to produce a similarity score. This score is compared to the ground truth distance (normalized) in the training loss function.
    }
    \label{fig:mtnn}
    \vspace{-10pt}
\end{figure*}

\subsection{Graph Convolutional Network (GCN) Similarity}
\label{subsec:simgnn}

Following~\cite{bai2019simgnn}, the computation of the \mt similarity has the following
steps: (1) a GNN transforms the node of each tree into a vector, this is called
node embedding; (2) a tree embedding is computed using attention-based aggregation;
(3) a joint embedding is obtained by comparing node and tree-level embeddings
to capture a comprehensive similarity between the \mts. (4) Finally, the
joint embedding is fed into a fully connected neural network to get the final
similarity score. Below, we provide more details on these steps.

\paragraph{Initialization} The inputs to our approach are \mts, stored as
adjacency matrices. The nodes of a \mt are weighted with the normalized~($[0,1]$)
function value. The ground truth distance is also normalized.

\paragraph{Node Embedding Computation} For the first step, we compute
a node embedding using a graph convolutional network
(GCN)\cite{kipf2016semi}, a type of GNN that updates the node features
by aggregating features from their neighbors. The formula for node level
updates in a GCN can be expressed as:
\vspace{-6pt}
\[
h_v^{(l+1)} = \sigma\left(\sum_{u \in \mathcal{N}(v)
\cup \{v\}}  \frac{1}{\sqrt{\hat{d_v}\hat{d_u}}} h_u^{(l)} W^{(l)} +
b^{(l)}\right),
\vspace{-6pt}
\]
where \(h_v^{(l+1)}\) is the feature vector of node \(v\) at layer \(l+1\);
\(\sigma\) is a non-linear activation function, such as ReLU$(x)=max(0,x)$;
\(\hat{d_v}\) is the degree of node $v$, plus 1 for self-loops; \(W^{(l)}\) is
the weight matrix at layer \(l\); and $b^{(l)}$ is the bias at layer \(l\). Given
a pair of \mts $\T_i,\T_j$, We obtain the node embedding $h_{iv}$ and
$h_{jv}$ for each node of $\T_i$ and $\T_j$ via the GCN.

\paragraph{Tree Embedding} For tree embeddings, we use attention-based aggregation. Attention-based aggregation is designed to assign weights to each
node based on a given similarity metric. $h_n \in \R^m$ is the embedding of $n$-th node, where $m$ is the size of embedding. The {\em global
context} $c\in \R^m$ is obtained as
%\vspace{-6pt}
\begin{equation}
\label{equ:c}
c = \tanh(\frac{1}{|V|}\sum_{n=1}^{|V|}h_n W_c),
%\vspace{-6pt}
\end{equation}
where $|V|$ is the number of nodes  and $W_c \in \R^{m\times m}$ is a learnable weight matrix. Next, each node should receive attention relative to this global context. The attention-weight embedding for a node is
%\vspace{-6pt}
\[
h^*_n= \mu(h_n^T c)h_n,
%\vspace{-6pt}
\]
where $\mu(\cdot)$ is the sigmoid function. The tree-level
embedding $H^*$ is aggregated from all re-weighted node embeddings, $h^*$. \add{$H^* = \sum_{n=1}^{|V|}  h^*$, where $|V|$ is the number of nodes in \mt.}

%Given the node-level feature $H \in \R^{|V|\times m}$,

%$H_i^*,H_j^*$, for MTs $\T_i,\T_j$, via

\paragraph{Joint Embedding Computation} For step three, the joint embedding is
obtained by comparing both tree-level and node-level embeddings.
We first describe how to compare the tree-level embedding using neural tensor
networks (NTNs) following~\cite{socher2013reasoning}:
\vspace{-6pt}
\[
D_{tree}(H_i^*,H_j^*)=\sigma(H_i^{*T}W_t^{[1:K]}H_j^* + V_w[H_i^* \cdot
H_j^*]+b_t),
\vspace{-6pt}
\]
where $D_{tree}$ is the tree-level vector that encodes similarity,
$\sigma(\cdot)$ is an activation function, $W_t^{[1:K]}\in \R^{m\times m \times
K}$ is a weight tensor, $[H_i^* \cdot H_j^*]$ is concatenation operation between
$H_i^*,H_j^*$, $V_w \in \R^{K\times 2m}$ is a weight vector, and $b_t\in \R^K$ is
a bias vector. Here, $K$ is a hyperparameter for the size of $D_{tree}$.  In summary,
the NTN provides a measure of similarity between two tree embeddings.

To assess node-level similarities, we compare the node embeddings from two
trees, \(\T_i\) and \(\T_j\). Letting \(N_i\) and \(N_j\) be the number of nodes
in \(\T_i\) and \(\T_j\),
respectively, this comparison yields an \(N_i \times N_j\) similarity matrix.
Specifically, the similarity matrix \(D_{node}\) is calculated as~\(D_{node} =
\sigma(H_i H_j^T)\), where \(H_i \in \R^{N_i\times m}\) and \(H_j \in
\R^{N_j\times m}\) are each the matrix of node embeddings for trees \(\T_i\) and
\(\T_j\). \(\sigma(\cdot)\) is an activation function that normalizes the scores into
the same range, \((0, 1)\). To address the discrepancy in node counts between trees,
we zero-pad the trees with fewer nodes \add{to match the node count of the trees with more nodes. Specifically, we append zero or null nodes to the end of the node list in the smaller trees. This approach ensures that the size difference between trees is accounted for in the similarity assessment without altering the original structure of the trees.}

\add{While zero-padding ensures that the node count between trees in each pair is the same, the overall size between different pairs can still differ. This is because the trees are padded only within each comparison pair to match the larger tree in that specific pair. However, the number of nodes in the larger tree can vary from one pair to another.} Following the methodology in~\cite{bai2019simgnn}, we convert this matrix into a
histogram \(\text{hist}(D_{node}) \in \R^B\), where \(B\) is a predefined number
of bins. This histogram representation standardizes the similarity matrix's
size, facilitating comparison across tree pairs.

\add{It is important to note that converting the similarity matrix \(D_{node}\) into a histogram standardizes the similarity matrix's size, facilitating comparison across tree pairs. However, this conversion can introduce distortions and artifacts because histograms are not continuous differential functions and do not support backpropagation.

To address this, we primarily rely on tree-level embedding comparisons to update model weights. The histogram-based node embedding comparisons are used to supplement the tree-level features, adding extra performance gains to our model. This approach ensures that the model benefits from the standardized representation of histograms while maintaining effective training through continuous tree-level embeddings.}

\paragraph{Final} In the final step, we combine the tree-level similarity~\(D_{tree}(H_i^*,
H_j^*)\) with the histogram \(\text{hist}(D_{node})\) as the joint embedding.
A fully connected neural network then processes this embedding to produce
the ultimate similarity score between the \mts pair in the $[0,1]$ range. This approach serves as the
baseline model in our study, with its performance detailed in
\secref{results}.

\subsection{Graph Isomorphism Network (GIN) Similarity}
\label{subsec:gin}

While our results with GCN similarity on \mt comparison are encouraging,
the variation in the number of nodes between two \mts plays a
significant role in distinguishing them. To emphasize the differences in node
count, we further improved the model by replacing the standard GCN with a
graph isomorphism network (GIN) for the node embedding computation.
As presented in~\cite{xu2018powerful}, GIN is adept at capturing
distinct graph structures. For instance, it can reproduce the
Weisfeiler--Lehman test for graph isomorphism.

The update rule for a GIN is:
\vspace{-6pt}
\[
    h^{(l+1)}_v = \text{MLP}^{(l)}\left((1 + \epsilon^{(l)}) \cdot h^{(l)}_v +
    \sum_{u \in \mathcal{N}(v)} h^{(l)}_u \right),
\vspace{-10pt}
\]
where \(h^{(l+1)}_v\) is node \(v\)'s feature vector at layer \(l+1\),
\(\text{MLP}^{(l)}\) denotes a multi-layer perceptron, and \(\epsilon^{(l)}\) is
a tunable parameter that balances the node's own features with its neighbors'.
The summation aggregates the features of node \(v\)'s neighbors, enhancing the
representation with local structural information.

By deploying a GIN, we enriched the model with a more complicated understanding
of node differences across \mts. In our work, we incorporated three GIN
layers to optimize node feature learning and report the performance in \secref{results}. The result shows that this approach improves performance
significantly over the GCN model.

However, it is crucial to note that GIN's capacity for graph isomorphism may not
fully capture the subtleties of MT comparisons. For instance, two \mts might be
structurally the same yet distinct in a topological sense due to differences in the function values of their nodes.
We extend GIN to a more customized
network for \mts to address this.  This is our final approach, called a merge
tree neural network (MTNN).

\section{Merge Tree Neural Networks (MTNN) Similarity}
\label{sec:mtnn}

As mentioned earlier, effectively capturing the topological characteristics of \mts
is crucial for learning their similarities using GNNs. We have
previously incorporated the function value for the nodes in a \mt \add{and applied a GIN to emphasize the node differences across \mts}.
However, we have not yet addressed the inclusion of another important
topological measure: the \textit{persistence} of features in \mts. To
integrate persistence, we use a
persistence-weighted adjacency matrix for tree embedding, which we detail in
\subsecref{persistenceedge}. Following that, in \subsecref{topoattn},
we describe how to apply this weighted matrix in our model effectively, and
we outline the learning process in \subsecref{learning}. An overview of the
model is provided in \figref{mtnn}.

\subsection{From Persistence to Edge Features}
\label{subsec:persistenceedge}

In the context of a \mt, $\mathcal{T}$, each edge connecting nodes $s$ and
$m$ represents \remove{a persistence pair} \add{the merging of critical points as the scalar value increases}, denoted as $(f(s),f(m))$, which we utilize as
an edge feature in our model. This approach, while useful, captures only incomplete
topological information, as certain topological features span across multiple
edges forming a path in the \mt, \add{For example, a persistence pair~$(b,d)$
represents the function values at the birth $(b)$ and death~$(d)$ of a feature,
where~$b$ is the function value at the local minimum and~$d$ is the function
value at the saddle point where this feature merges with another.}

\begin{figure}[t]
    \centering % avoid the use of \begin{center}...\end{center} and use \centering instead (more compact)
    \includegraphics[width=\linewidth]{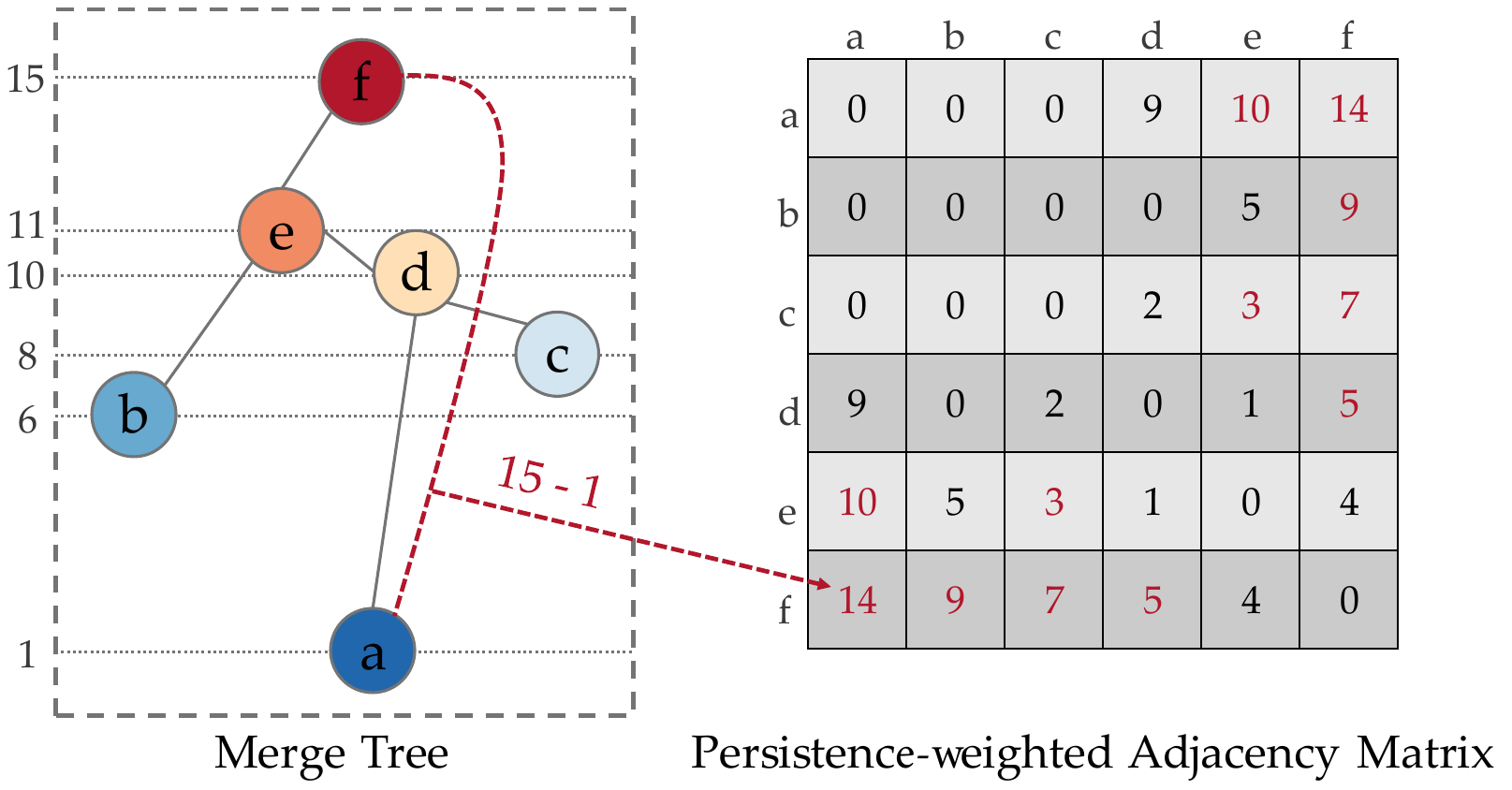} %edge}
    %\vspace{-18pt}
    \caption{An example of our persistence-weighted adjacency matrix. \remove{All edges are weighted by persistence.} \add{Each entry in the matrix represents the function value difference between connected nodes in the merge tree. Values highlighted in red denote entries for nodes that are not directly connected in the merge tree but are included to capture the persistence pairs. This enhancement allows for the inclusion of broader topological connections, ensuring the full topological characteristics of the merge tree are represented.}
    }
    \label{fig:persistence_edge}
    \vspace{-10pt}
\end{figure}

To summarize the full topological characteristics of a \mt, we introduce
a persistence-weighted edge adjacency matrix $\hat{E}$. This matrix is
constructed by considering the function value differences between connected
nodes and their connections to neighboring nodes.
\add{Specifically, each entry in the matrix represents the function value difference between nodes, recording all paths.} This enhancement allows us to
extend our analysis from direct pairs like $(f(s),f(m))$ to broader connections,
capturing $(f(s),f(\mathcal{D}(m)) \cup f(m))$, which includes node~$m$ and its
descendants, $\mathcal{D}(m)$. \add{Therefore, the complete topological information is encoded, as the persistence pair has been added.} \figref{persistence_edge} illustrates how we accomplish this construction.

To integrate this persistence-weighted matrix, we tested two methods: (1)
Incorporating it into GNN architectures like GIN or GCN; (2) Utilizing it in
an attention-based aggregation to map node embeddings to tree embeddings.

Option (1) introduces modified edges (referred to as  "pseudo" edges) to include
\textit{persistence} information. Consequently, this approach updates node
features by utilizing neighbor features via the adjacency matrix. However, this
modification changes the original \mt structure in the message updates
function of GNNs due to the introduction of pseudo edges.

To preserve the original \mt structure while incorporating
\textit{persistence} information, we choose method (2). This approach leverages
the persistence-weighted matrix in an attention-based aggregation process,
mapping node embeddings to tree embeddings \add{(Sec.~\ref{subsec:topoattn})}. Here, node re-weighting is informed
by their persistence, but not exclusively so. We train the weight matrix to
consider both the persistence information and the overall \mt distance \add{(Sec.~\ref{subsec:learning})},
ensuring a balanced integration of topological features.

\subsection{Topological Attention}
\label{subsec:topoattn}

This section outlines the design of topological attention, leveraging the
persistence-weighted adjacency matrix $\hat{E}$.

Building on the attention-based aggregation discussed in
\subsecref{simgnn}'s Tree Embedding, we reformulate the
{\em global context} vector $c$\add{, replacing Eq.\ref{equ:c}}. We integrate the topological information as:
\vspace{-10pt}
$$
Norm = \sum_{k=1}^{|V|} \sum_{l \in \mathcal{N}(k)}\hat{e}_{kl},
\vspace{-10pt}
$$
$$
c = \tanh\left(\frac{1}{|V|}\sum_{n=1}^{|V|}\left(\frac{\sum_{u\in
\mathcal{N}(n)} \hat{e}_{un}}{Norm}\right)h_n W_c\right),
\vspace{-6pt}
$$
where $u\in \mathcal{N}(n)$ is a neighbor node of $n$, $\hat{e}_{un} \in
\hat{E}$ is the persistence-weighted edge feature of $u$ and $n$, $h_n$ is the
embedding of $n$-th node,~$|V|$ is the number of nodes, and $W_c$ is a learnable weight matrix . To further break down the
the formula above, we have the local weighting factor for each
node $v$, the sum \(\sum_{u\in \mathcal{N}(v)} \hat{e}_{uv}\) computes the total
edge weight connected to \(v\).
By dividing this sum by the normalization term  \( Norm \), we normalize the local
weighting with respect to the total edge weights in the tree, ensuring the scale
of the features remains consistent. Then, we use the same calculation as the
previous attention-based aggregation. \add{Note that $c$ is used to compute $h^{*}$ which in turn is used to compute the term, $H^{*}$, used in training.}

\subsection{MTNN Learning}
\label{subsec:learning}

Training of the MTNN uses a Siamese network architecture,  utilizing GINs as
encoders to transform
input \mts into node embeddings. We generate joint embeddings by combining node
and tree embeddings, where the tree embedding is derived using a topological
attention-based aggregation. This aggregation reweights nodes according to their
topological features. Subsequently, we employ an MLP-based regression network to
map the joint embedding to the ground truth similarity score
between the \mts. The model is trained to minimize the Mean Squared Error (MSE)
loss, defined as:
\vspace{-6pt}
$$
    L = \frac{1}{D}\sum_{i,j \in D}(\text{MLP}(H_{ij}^*) - s_{ij})^2
\vspace{-6pt}
$$
Here, $\text{MLP}$ denotes the MLP-based regression network, and $D$ is the set
of all training \mts pairs, $H_{ij}^*$ is the joint embedding and  $s_{ij}$ is the ground truth MT distance.

%% file: body/results.tex
\begin{table}[htb]
    \caption{Number of \mts (MTs), \add{the simplification threshold, $\tau$, employed (same as used in previous work)}, and their range in node counts after
    simplification.}
    \label{tab:dataset}
    \vspace{-18pt}
    \begin{center}
        \begin{tabular}{@{}llll@{}}
            \toprule
            Dataset         & \# of MTs & $\tau$ & \# of Nodes \\
            \midrule
            {MT2k}          & 2000  &  0.1  & [8,191] \\
            {Corner Flow}   & 1500  &  0.2  & [24,30] \\
            {Heated Flow}   & 2000  &  0.06  & [12,27] \\
            {Vortex Street} & 1000  &  0.05  & [56,62] \\
            {TOSCA}      & 400      &  0.01  & [12,78] \\
            \bottomrule
        \end{tabular}%
    \end{center}
     \vspace{-24pt}
\end{table}

\begin{figure}[bt]
    \centering % avoid the use of \begin{center}...\end{center} and use \centering instead (more compact)
    \includegraphics[width=\linewidth]{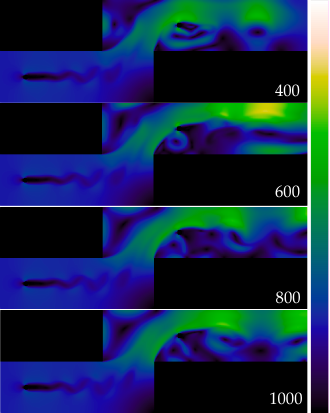} %edge}
    \vspace{-18pt}
    \caption{Time steps of the 2D viscous Corner Flow simulation dataset.
    }
    \label{fig:cf_exp}
    \vspace{-14pt}
\end{figure}

\section{Results}
\label{sec:results}
While our proposed approach can be generalized to different merge tree distances, we have chosen the state-of-the-art distance metric from~\cite{yan2022geometry} as our ground truth. This metric is noted for its efficient computation and is accompanied by an accessible open-source implementation. As mentioned, this distance is normalized to fall within the $[0,1]$ range to coincide with a similarity score. To compare the quality of our MTNN similarity in reproducing \cite{yan2022geometry}, we use Mean Squared Error (MSE) as our evaluation metric.

\subsection{Datasets}
\label{subsec:datasets}

We evaluate MTNN on five datasets: MT2k, Corner Flow, Heated Flow, Vortex
Street, and TOSCA. All datasets, except MT2k, were chosen because they have
previously been used in \mt distance
research~\cite{yan2022geometry,sridharamurthy2018edit}. Interestingly enough,
three~\cite{yan2022geometry} are from a similar domain: 2D flow simulations.
This allows us to test the general applicability of a trained MTNN model
across the same domain.
Following the previous work, we de-noise each \mt for a pre-determined
threshold $\tau$ as follows: given a node pair $(s,m)$,
where $s$ is a local minimum and~$m$ is its emerging saddle point, the
persistence is computed for each pair,~$p = f(m) - f(s)$, where
$f(s)$ and $f(m)$ are the function values at node $s$ and $m$, respectively.  If $p$ is less
than $\tau$, then node $s$ and its connecting edge are removed with
the children of $s$ being directly connected to~$m$. Nodes are processed in reverse order of persistence. This results in the \mt
having fewer nodes and edges, with only the significant features remaining. More
information on \mt simplification can be found
in~\cite{edelsbrunner2010computational,yan2022geometry,sridharamurthy2018edit}.
The
number of \mts and range of node counts after simplification for each are
summarized in \tabref{dataset}.
%\brittany{how did we pick $\tau$? was this fixed for each data set? if so, it should be in the table. either way, we need to describe.}

As mentioned, all \mts used are {\em join trees.}, although the approach could
have also easily used {\em split trees}. \add{The comparison uses a simplified tree based on the same parameters of previous work.} Finally, we used the standard random
split, $80\%$ and $20\%$, of all the \mts as training and testing sets for each
dataset.

\paragraph{MT2k} This dataset is 2000 synthetic 3D point clouds with two
distinct classes. The first features three noisy tori, and the second class
contains three noisy tori plus one noisy sphere. Each point cloud is constructed
by synthetically sampling 100 points from the respective geometric shapes.
Random noise is added during the sampling to ensure the uniqueness of each data
point. Corresponding \mts are constructed for each point cloud to represent
their topological features. We apply a persistence threshold~($\tau=0.1$) in the
\mt simplification process.

\paragraph{Corner Flow} This dataset is 1500 time steps from a simulation of 2D
viscous flow around two cylinders~\cite{gerrisflowsolver,BaezaRojo19SciVisa}.
Following~\cite {yan2022geometry}, we generate a set of \mts from the vertical
component of the velocity vector fields.
We also use the same persistence threshold~($\tau=0.2$) for \mt simplification
as the previous work. See \figref{cf_exp}.

\begin{figure}[bt]
    \centering % avoid the use of \begin{center}...\end{center} and use \centering instead (more compact)
    \includegraphics[width=\linewidth]{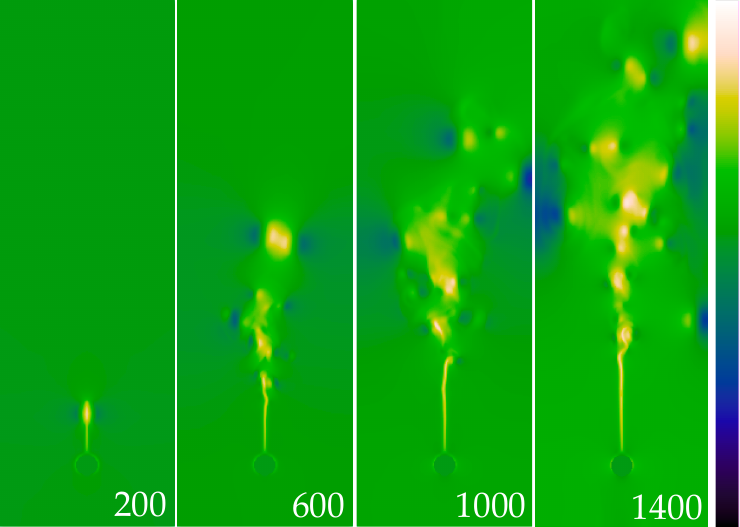} %edge}
    \vspace{-18pt}
    \caption{Time steps of the 2D Heated Flow simulation dataset.
    }
    \label{fig:hf_exp}
    \vspace{-10pt}
\end{figure}

\paragraph{Heated Flow} This dataset is from a simulation of the 2D flow created
by a heated cylinder using a Boussinesq
Approximation~\cite{gerrisflowsolver,Guenther17}. Following~\cite
{yan2022geometry}, we convert each time instance of the flow into a scalar field
using the magnitude of the velocity vector. The persistence simplification
threshold used here is the same as the previous work~($\tau=0.06$). See \figref{hf_exp}.

\begin{figure}[tb]
    \centering % avoid the use of \begin{center}...\end{center} and use \centering instead (more compact)
    \includegraphics[width=\linewidth]{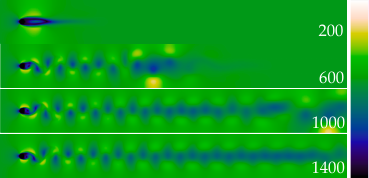} %edge}
    \vspace{-18pt}
    \caption{Time steps of the 2D von-Kármán Vortex Street dataset.
    }
    \label{fig:vs_exp}
    \vspace{-14pt}
\end{figure}

\paragraph{Vortex Street} This is an ensemble of 2D regular
grids~\cite{gerrisflowsolver,Guenther17}, each with a scalar function defined on
the vertices to represent flow turbulence behind a wing, creating a 2D
von-Kármán vortex street.  Following~\cite {yan2022geometry}, we use the
velocity magnitude field to generate \mts and apply a persistence simplification
with the threshold ($\tau=0.05$) to the \mts. See \figref{vs_exp}.

\begin{figure}[tb]
   \centering % avoid the use of \begin{center}...\end{center} and use \centering instead (more compact)
   \includegraphics[width=0.75\linewidth]{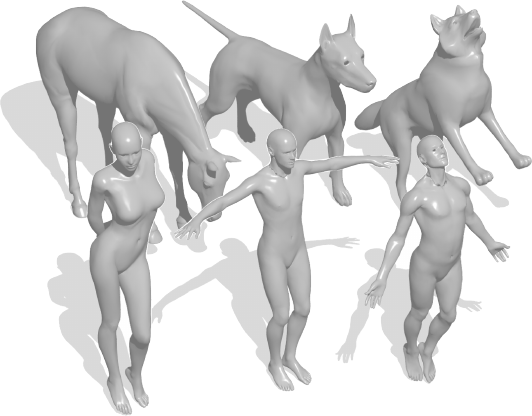} %edge}
   \vspace{-8pt}
   \caption{Example 3D models from the TOSCA dataset.
   }
   \label{fig:3d_exp}
\end{figure}

\paragraph{TOSCA} This dataset~\cite{bronstein2008numerical} contains a collection of different, non-rigid shapes of animals and humans.
Following~\cite{sridharamurthy2018edit}, we compute the average geodesic
distance field on the surface mesh. Like the previous work, persistence
simplification is used with the threshold~($\tau=0.01$). See \figref{3d_exp}.

\subsection{Implementation Details}
\label{subsec:details}

Merge trees are computed using TTK~\cite{tierny2017topology} and
Paraview~\cite{squillacote2007paraview} is used for dataset visualization.  Our
implementation utilizes
\href{https://pytorch.org/}{PyTorch}~\cite{paszke2019pytorch}.
We employ a three-layer GIN~\cite{xu2018powerful} as the encoder network with ReLU
activation function with all weights initialized randomly. The output dimensions
for the 1st, 2nd, and 3rd GIN layers are 64, 32, and 16, respectively. In the
NTN layer, we specify $K=16$. Following the approach in~\cite{bai2019simgnn}, we
use 16 histogram bins for pairwise node embedding comparisons.  For training, we
choose the Adam optimizer~\cite{kingma2014adam}, setting the learning rate to
$0.001$ and the weight decay to $0.0005$. The model is trained with a batch size
of $128$ over 100 epochs. All datasets are converted into standard dataloaders
compatible with
\href{https://pytorch-geometric.readthedocs.io/en/stable/}{PyTorch Geometric
(PyG)}~\cite{fey2019fast} for GNN processing. To present each technique in the
best light, we computed all timings for our ground truth
distance~\cite{yan2022geometry} on a machine with a 16-Core (8P/8E) Intel
I9-12900K CPU @ 5/4 GHz with 32 GB memory, as the approach benefits most from
multiple cores.  MTNN timings are computed on a machine equipped with a six-core
Intel i7-6800K CPU @ 3.50GHz, 68GB of memory, and an Nvidia 3090Ti GPU, since it
benefits most from a better GPU.

\begin{figure}[t]
\centering
\includegraphics[width=\linewidth]{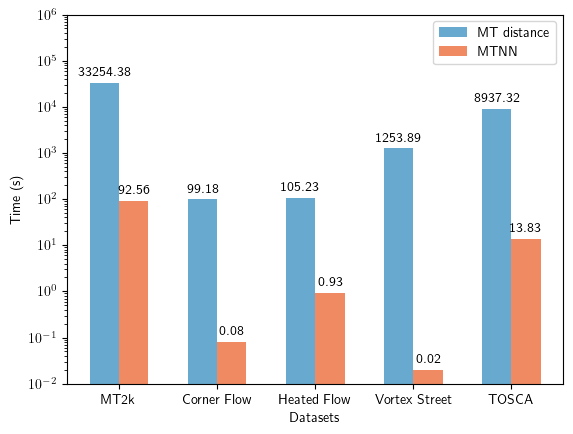}
\vspace{-18pt}
\caption{Comparison of computational times for \mt distance~\cite{yan2022geometry} and MTNN across different datasets. Times are presented in \textbf{log scale}. (Runtime for \mt distance is computed with 16 cores)}
\label{fig:runtime}
\end{figure}

\subsection{Evaluation Results}
\label{subsec:result}

We evaluate the proposed approach from two perspectives: effectiveness and
efficiency. For effectiveness, we compute the distance between the predicted
similarity and our ground truth distance~\cite{yan2022geometry} using mean
squared error (MSE), the standard for learned graph similarity validation. For
efficiency, we record the wall time needed to compute the full pairwise
similarity matrix for the testing set.

\paragraph{Effectiveness} \tabref{ablation} provides quality results for our
machine learning approaches. This includes the initial GCN, improved GIN, and
our final MTNN model with topological attention. This study focused on models
trained and tested on the same (split) dataset. For readability, we have scaled
each by $10^3$. Since both our ground truth and similarity output are in the
$[0,1]$ range, the error is $< 0.1\%$. Therefore the MSE between our approach
and the ground truth is extremely low. The table also shows that the GIN model
consistently outperforms our initial GCN model.  Finally, our MTNN model in this
scenario is comparable to or better than our GIN formulation. Our MTNN approach
outperforms GIN in cases of higher GIN error, but has diminishing returns as the
error lowers.
\begin{table}[tb]
    \caption{MSE error for test datasets for GCN, GIN, and our MTNN networks.
        Each dataset was trained on a standard split of the \textbf{source data}.
        All results are scaled by $10^{3}$ for readability, as our errors are
        extremely low.  This means that all trained networks reproduce the
        similarity very close to the ground truth. Our MTNN network improves the
        quality of the reproduction, but not in cases where GIN error is already
        extremely low.}
            \vspace{-12pt}
    \label{tab:ablation}
    \begin{center}
        \begin{tabular}{@{}llll@{}}
            \toprule
            Dataset         & GCN           & GIN                    & MTNN                   \\
            \midrule
            {MT2k}          & 0.53819   & 0.18996                & \textbf{0.10408} \\
            {Corner Flow}   & 65.23393  &  \textbf{0.00145} & 0.00312                \\
            {Heated Flow}   & 36.36125  & 0.00468                &  \textbf{0.00263} \\
            {Vortex Street} & 470.82582 &  \textbf{0.00008} & 0.00018                \\
            {TOSCA}      & 12.60986  & 0.01542                & \textbf{0.00986} \\
            \bottomrule
        \end{tabular}%
    \end{center}
        \vspace{-18pt}
\end{table}

\tabref{generalized} explores the generalizability of our models.  As before,
all error results are scaled by $10^3$ for readability. First, we tested the
quality of our similarity measure in a scenario where the model is trained on
\mts from one data type but is applied to other types.  For this, we used a
model trained on the synthetic MT2k dataset.  The table shows that the error is
quite low even though we use \mts from a 3D point cloud on very different data
(i.e., 2D flows and 3D shapes). MT2k is also a fairly limited and constrained
data source.  The quality of these results shows that the MTNN model has the
potential to be generally applied.
\begin{table}[tb]
\caption{MSE error for our datasets on models trained on other data.  On the
        left is the error of applying the MT2k-trained model.  On the right is a new
        model trained with a mix of Corner Flow, Heated Flow, and Vortex Street. Our
        topological attention is most effective in this scenario, leading to the
        lowest error values for each model (bold).  Our synthetic MT2k performs best
        on the 3D shape dataset, while the model trained on the 3 separate flow
        datasets is best for the flow data (underlined).  All results are scaled by
        $10^{3}$ for readability, so all models return a low error with most $<
        0.1\%$.}
    \label{tab:generalized}
    \vspace{-6pt}
    \resizebox{\columnwidth}{!}{%
        \begin{tabular}{@{}l|lll|lll@{}}
            \toprule
            &
            \multicolumn{3}{c|}{Trained on MT2k} &
            \multicolumn{3}{c}{Trained on CF + HF + VS} \\
            &
            {GCN} & {GIN} & MTNN &
            {GCN} & {GIN} & MTNN \\
            \midrule
            {Corner Flow} &
            {83.93713} &
            {0.03507} &
            \textbf{0.01795} &
            {17.20086} &
            {1.60534} &
            {\ul \textbf{0.00862}} \\
            {Heated Flow} &
            {57.93813} &
            {0.16013} &
            \textbf{0.00501} &
            {32.83913} &
            {0.23754} &
            {\ul \textbf{0.00573}} \\
            {Vortex Street} &
            {541.03814} &
            {113.24254} &
            \textbf{3.78652} &
            {178.93201} &
            {63.92713} &
            {\ul \textbf{0.17532}} \\
            {TOSCA} &
            {27.9381} &
            {3.17674} &
            {\ul \textbf{0.2016}} &
            {37.8729} &
            {8.03914} &
            \textbf{1.63408} \\
            \bottomrule
        \end{tabular}%
        }
\end{table}

\begin{figure}[tb]
    \centering % avoid the use of \begin{center}...\end{center} and use \centering instead (more compact)
    \includegraphics[width=\linewidth]{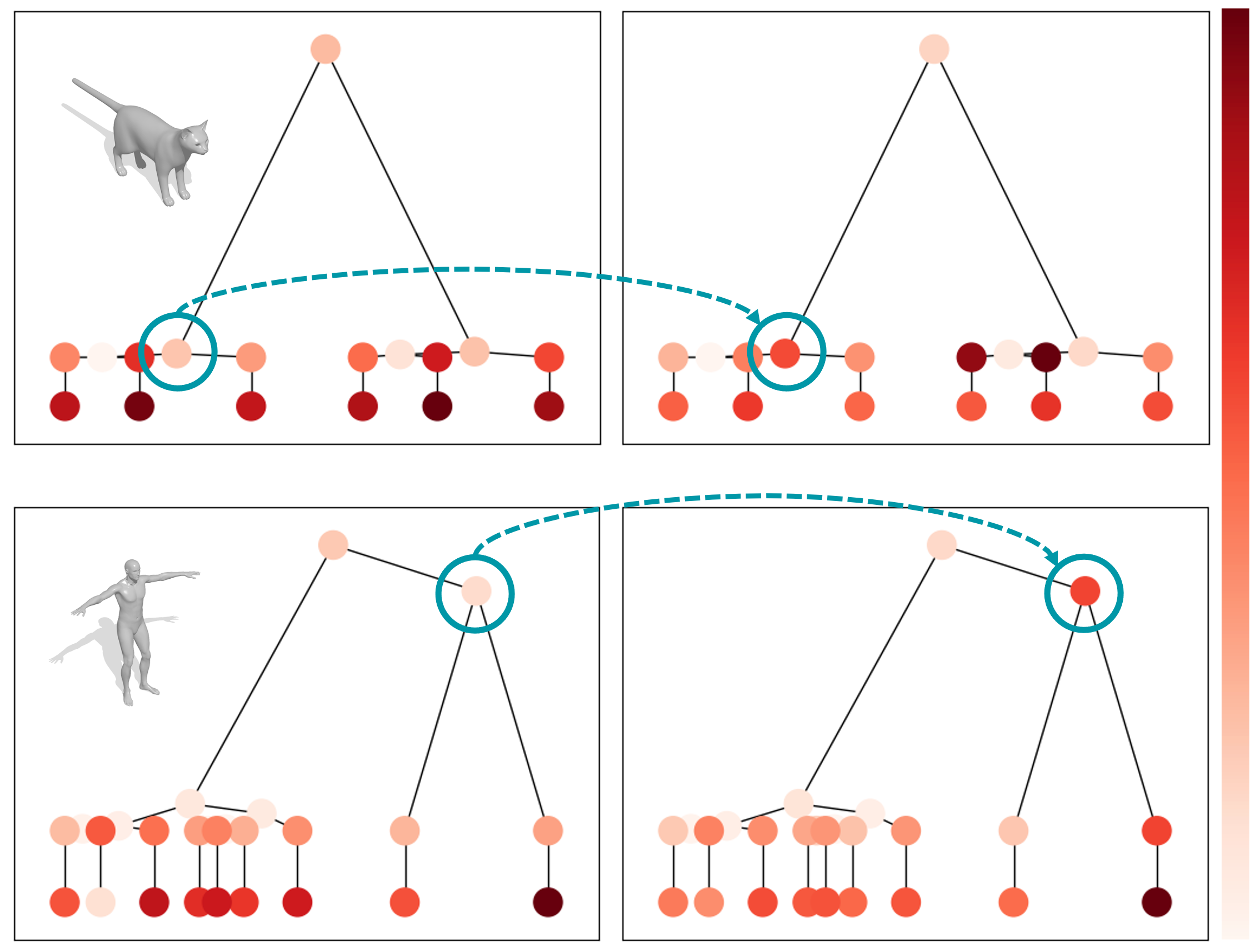} %edge}
    \vspace{-18pt}
    \caption{Topological attention on two merge trees from the TOSCA dataset. \add{Each row displays the merge tree visualized by the relative importance, with example inserts within the merge trees}. Left: node importance without topological attention, using the GIN model. The node colors indicate their relative importance when comparing the two examples \add{(the darker with red, the higher the weight)}.  Right: node importance with topological attention using the MTNN model. The GIN model already emphasizes structural differences between the two merge trees. But, the introduction of topological attention further emphasizes these differences, re-weighing the nodes in the high persistence feature highlighted in cyan.
    }
    \label{fig:tosca_attn}
    %\vspace{-20pt}
\end{figure}

\begin{figure}[tb]
    \centering % avoid the use of \begin{center}...\end{center} and use \centering instead (more compact)
    \includegraphics[width=\linewidth]{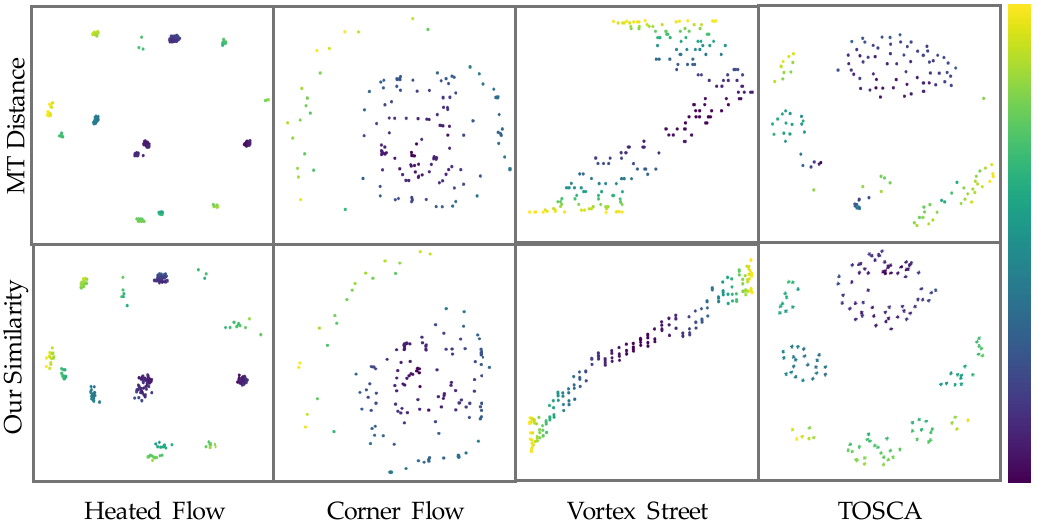} %edge}
    \vspace{-18pt}
    \caption{Results from multidimensional scaling (MDS) analysis. Two MDS maps represent each dataset. On the top are the maps created using a similarity matrix based on the ground truth \mt (MT) distance~\cite{yan2022geometry}. On the bottom are maps generated from a similarity matrix of MTNN. The coloring indicates the mean distance, showing how close or far points are from each other in the MDS space. Comparing the two maps shows similar patterns in colors and point arrangements. This similarity indicates that MTNN effectively learns and reflects the relationships between points.
    }
    \label{fig:mds_all}
\end{figure}

Next we tested how a model trained on a mix of {\em like} data performs.  In
this scenario, we trained on the combination of the Corner Flow, Heated Flow,
and Vortex Street training sets, each sampled equally by 800.  This model was then
applied to our test sets. As is shown, the error is very low for our vector
field datasets. This means that our model seems to generalize well across fields
from this domain.  In addition, it performs quite well on the 3D shape dataset
but not as well as the MT2k-trained model. This makes intuitive sense since
TOSCA and MT2k are both geometric 3D datasets. Therefore, with these results, we
can postulate that this model can be generalized with low error. However, for the
lowest error possible, it is best to train on data in the same domain but not
necessarily from the same source (e.g., simulation).

\paragraph{Efficiency}
As stated in the beginning, this work aims to create an approach that can compute \mt
similarity, not only faster than the state-of-the-art, but also at speeds that
could potentially support real-time analysis.  \figref{runtime} provides
our runtimes compared to the fast \mt distance calculation of
\cite{yan2022geometry}. Also, as detailed in the introduction to this section,
we chose machines from our available hardware on which each technique
performs best (i.e., more cores vs.\ a better GPU). Timings are based on the total
time to compute all pairwise distances for each test dataset. Note that the
times are plotted in log scale. As this figure shows, not only are we
significantly faster, we are orders of magnitude faster ([2-5], 3 median). In
addition, the time to compute similarity for a single pair of \mts is so small
that it can be considered negligible.  Therefore MTNNs can potentially support
interactive queries.

\add{
\begin{table}[tb]
    \centering
    \caption{Training times for different datasets (hours).}
    \vspace{-14pt}
   \begin{center}
\resizebox{\columnwidth}{!}{%
    \begin{tabular}{@{}lcccccc@{}}
        \toprule
        Dataset & MT2k & Corner Flow & Heated Flow & Vortex Street & TOSCA \\
        \midrule
        Training Time (h) & 4.28 & 2.86 & 3.83 & 1.93 & 3.95 \\
        \bottomrule
    \end{tabular}
    }
    \end{center}
    %\vspace{-24pt}
    \label{tab:train}
\end{table}

\tabref{train} shows the training times for each dataset using our approach.
This demonstrates the efficiency of our method, which is comparable to other GNN
methods. Note that these timings are insignificant when compared to the time
needed to compute the ground truth distances for training. Additionally, our
model is potentially generalizable across different datasets, as shown in
\tabref{generalized}, possibly mitigating training~costs.}

\subsection{Further Analysis and Visualization}
To further assess our method, we visualize the effect our topological attention
mechanism has on merge trees. Fig.~\ref{fig:tosca_attn} gives an example pair
from TOSCA. Attention weights on \mt nodes are displayed in red, with the
intensity indicating the weight magnitude: the darker the red, the higher the weight.
This figure shows node importance with~(left) and without (right) our topological attention mechanism.

As highlighted in cyan, the high persistence feature in the human model increases in weight due to our attention approach, which makes intuitive sense. \add{It is interesting to see that the saddle point associated with the global maximum obtains more weight after applying topological attention in both the human and cat models. For the cat model, the left saddle point associated with the global maximum becomes more important, and the overall leaf node becomes less important. In the human model, the right saddle point gains more importance, and only the leaf node of the left subtree becomes less significant. These observations indicate that topological attention is highly data-dependent. This dependency demonstrates the usefulness of topological attention, as it adapts to the unique characteristics of each dataset, highlighting critical features that might otherwise be overlooked.}

The extremely low error, as detailed in Table \ref{tab:ablation}, demonstrates that our method accurately learns pairwise similarities, motivating us to explore how well it reproduces the entire similarity matrix. We applied multidimensional scaling (MDS)~\cite{carroll1998multidimensional} to the similarity matrix constructed from our ground truth \mt distance~\cite{yan2022geometry} and the matrix formed from MTNN. MDS maps these matrices into 2D space, which we can visualize side-by-side. We assigned colors to each point in the MDS plots based on their average distance to other points. If MTNN captures the similarity of our ground truth, we would expect to see similar color patterns and point arrangements.  While there are some differences in the visualization, on the whole, as demonstrated in Fig.~\ref{fig:mds_all}, the structures and patterns are indeed similar.

%% file: body/conclusion.tex
\section{Conclusion}
\label{sec:conclusion}
In this study, we addressed the challenge of topological comparison, focusing
specifically on merge trees. By reconceptualizing this problem as a
{\em{learning}} task, we introduced merge tree neural network (MTNN) that employs graph neural networks (GNNs) for comparisons. Our approach is both fast and precise leading to comparisons that are orders of magnitude faster and extremely low in added error when compared to the state-of-the-art.

 \add{MTNN is the first deep learning model specifically designed for \mts comparison, combining modified GNNs with novel topological attention to jointly learn the distance metric and topological properties of \mts. MTNN introduces several key technical advancements over standard GNNs like GCN: (1) MTNN addresses node count discrepancies in \mts comparison by employing GIN for node embedding computation, emphasizing the differences in node counts; (2) MTNN enhances the representation of topological information by initializing node features with function values derived from the \mt; (3) MTNN incorporates topological attention by explicitly integrating persistence information into the aggregation function, enabling the model to capture and utilize the intrinsic topological properties of \mts more effectively.}

%Our approach used a trained model to transform merge trees into vector spaces, significantly enhancing the speed of comparison against traditional methods.

%Our approach's unique advantage is its ability to learn similarities without requiring predefined classes for merge trees. This shift to similarity learning broadens the potential applications for merge tree analysis, emphasizing the centrality of similarity in their analytical use.
While our method represents a significant advancement in merge tree comparisons, it also presents challenges, such as the need for data to train. However, our experiments demonstrate the model's potential for generalization, which can mitigate this issue in its practical application. In addition, our test datasets only needed training sets of size in the hundreds or low thousands, which is relatively low for accurately trained models.

Future directions for our work include applying our merge tree comparisons to various analytical tasks in scalar field analysis, such as fast topological clustering or error assessment in approximated scalar functions. \add{A key focus of our future work will be exploring more applications and datasets that could benefit from the use of merge trees.} Our work opens up a new direction at the intersection of machine learning and topological data analysis. We believe that our work will encourage further investigations and developments in this emerging field, driving advancements in both theoretical understanding and practical applications.

%% file: body/supplemental.tex
%\section*{Supplemental Materials}
%\label{sec:supplemental_materials}

%Python implementation url here. 